\documentclass[11pt]{article}

\usepackage[margin=1in]{geometry}
\usepackage{amsmath, amssymb}
\usepackage{booktabs}
\usepackage{graphicx}
\usepackage{hyperref}
\usepackage[numbers,sort&compress]{natbib}
\usepackage{xcolor}
\usepackage{tikz}
\usetikzlibrary{positioning,arrows.meta,fit,backgrounds}

\title{Student-Guided Teacher Distillation for Efficient LLM Task Routing:\\ \large Positioning Against Jev-Style System-1 Classifiers}

\author{
  Haifeng Wu \quad Srinivasan Manoharan \quad Jian Wan \quad Fangbo Tu \quad Junhua Zhao \quad Xin Chen \\[4pt]
  PayPal \\
  \texttt{\{haifwu, srinivmanoharan, jwan, fatu, zhahua, xchen7\}@paypal.com}
}

\date{\today}

\begin{document}

\maketitle

\begin{abstract}
Large zero-shot classifiers are useful for routing user requests to specialized LLM tasks, but scoring every request against a large candidate set is expensive: a zero-shot natural language inference (NLI) classifier must evaluate one premise-hypothesis pair per candidate label, so its cost scales linearly with the size of the taxonomy. We study a student-guided teacher distillation pipeline for a fixed taxonomy of 60 LLM task categories, in which a compact ModernBERT classifier predicts the full category distribution in a single forward pass and retrieves a small top-$k$ candidate set, and a larger DeBERTa-v3 zero-shot NLI classifier then reranks only those candidates rather than all 60 labels; the resulting teacher labels iteratively improve the student, which in turn produces sharper candidate sets for the next labeling round. Unlike generic embedding retrieval or the clustering-derived shortlists used in extreme multi-label classification, the candidate generator here is trained end-to-end on the target taxonomy and is the same model that serves production traffic, distinguishing it from LLM-routing work that routes between candidate \emph{models} rather than into a fixed semantic \emph{task} taxonomy, and from concurrent ``System 1'' encoder-classifier proposals such as TypeSafe AI's Jev~\citep{typesafe2026jev} and the open-source Laya~\citep{laya2026} project, whose training methodology is undocumented or RL-based. Our best student checkpoint reaches 77.5\% teacher agreement on a 200-example evaluation set (the deployed checkpoint, v4, reaches 76.5\%), and preliminary coverage measurements on two convenience samples show Coverage@16 of 91--100\%, suggesting top-$k$ candidate sets retain most of the teacher's decision-relevant information. We further show that truncated top-$k$ teacher scores should not be treated as full 60-class soft targets for KL distillation, since zeroing the untruncated classes destroys the ``dark knowledge'' that soft-label distillation depends on and introduces a systematic bias rather than a harmless sparse approximation. A complete evaluation of the student-guided iterative method --- including candidate coverage at multiple $k$ on a certified held-out set, an embedding-retrieval baseline, and a larger human-reviewed test set --- remains in progress and is scoped explicitly in Section~\ref{sec:required-eval}.
\end{abstract}


\section{Introduction}

LLM applications increasingly rely on routers that classify a user request before selecting a downstream agent, tool, or model. A router must be fast because it runs on every request, but it must also distinguish between fine-grained and semantically similar tasks such as debugging, code review, deployment, documentation, and pull-request management.

Zero-shot natural language inference models provide a convenient way to build such routers without collecting a large manually labeled dataset. However, the teacher must evaluate the input against every candidate label as a separate premise-hypothesis pair. With 60 labels and hundreds of thousands of requests, this repeated pairwise computation becomes the dominant cost of dataset creation.

Knowledge distillation can produce a fast student that predicts all 60 classes in one forward pass. The central observation of this work is that the student can also be used \emph{before} teacher inference: its probability vector provides a task-aware candidate ranking. The teacher then only needs to rerank a small top-$k$ subset.

\begin{figure}[!ht]
\centering
\begin{tikzpicture}[
  node/.style={draw, rounded corners, minimum height=0.9cm, align=center, font=\small, inner sep=6pt},
  ->, >=Latex, thick
]
\node[node] (student) {student$_t$};
\node[node, right=1.3cm of student] (topk) {top-$k$\\candidate labels};
\node[node, right=1.3cm of topk] (teacher) {teacher\\reranking};
\node[node, right=1.3cm of teacher] (sup) {hard/soft\\supervision};
\node[node, below=1.1cm of student] (student2) {student$_{t+1}$};

\draw (student) -- (topk);
\draw (topk) -- (teacher);
\draw (teacher) -- (sup);
\draw (sup.south) |- (student2.east);
\draw (student2) -- (student) node[midway, left, font=\scriptsize] {retrain};
\end{tikzpicture}
\caption{The iterative student-guided labeling loop. The student's top-$k$ candidates bound what the teacher must score; the resulting hard/soft labels retrain the student for the next round, which can then attempt a smaller $k$ once coverage at the current $k$ is confirmed.}
\label{fig:loop}
\end{figure}
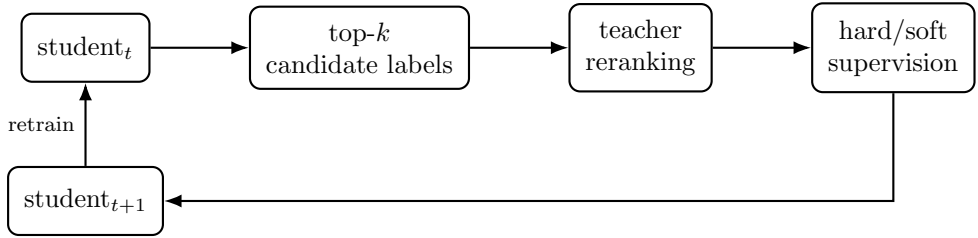

We investigate the iterative process illustrated in Figure~\ref{fig:loop}: the student proposes a top-$k$ candidate set, the teacher reranks only those candidates, and the resulting hard/soft labels retrain the student for the next round.

The goal is not to replace the teacher's decision rule with semantic similarity. The goal is to reduce the teacher's labeling cost while preserving the teacher's final classification behavior.

\paragraph{Why this is not just faster NLI.} A concurrent and unrelated line of industry work makes a related efficiency argument at a different layer of the stack. TypeSafe AI's ``System One models,'' introduced with the Jev model~\citep{typesafe2026jev}, propose training a single model to emit calibrated, type-safe structured decisions directly, positioned as a fast ``System 1'' layer beneath slower, deliberative ``System 2'' LLM reasoning. As of this writing, no technical report, architecture description, or training methodology for Jev or its ``RLCD'' (Reinforcement Learning for Calibrated Decisions) training procedure has been published beyond marketing benchmarks on TypeSafe's own site; independent community discussion has not established what, mechanistically, distinguishes it from constrained decoding over an existing model~\citep{typesafe2026jev}. We note this only as motivating context for why fast, calibrated, structured decision-making is an active industry concern, and flag for the record that TypeSafe's ``RLCD'' acronym is an unrelated coincidental reuse of the name from \citet{yang2023rlcd}'s earlier and unrelated \emph{Reinforcement Learning from Contrastive Distillation} alignment method. The System~1/System~2 framing itself is also not new to either work: it has previously been used to describe distilling slow chain-of-thought reasoning into fast direct generation~\citep{yu2024distilling}. A further, independently confusing detail is that an open-source project, Laya~\citep{laya2026}, has since positioned itself as a direct Jev alternative --- also built on a ModernBERT-family encoder (ModernBERT-large and mmBERT-base) for single-forward-pass typed decisions, and also trained with what it calls ``RLCD,'' though there it denotes reinforcement learning against strictly proper scoring rules rather than TypeSafe's ``Reinforcement Learning for Calibrated Decisions'' or \citet{yang2023rlcd}'s \emph{Reinforcement Learning from Contrastive Distillation}: three unrelated methods now share one acronym. We cite Laya only as further evidence that compact encoder classifiers are being actively explored as fast alternatives to Jev-style System~1 models --- it is, like Jev, a README-level project with no accompanying technical report or peer-reviewed evaluation at time of writing, and its training method (policy-gradient optimization against a scoring rule, over general typed-decision tasks) and evaluation focus (open-ended typed decisions across 100+ languages) are unrelated to the teacher-distillation, fixed-60-class-taxonomy routing problem this paper studies. Our contribution differs from this entire family in scope and in what is actually demonstrated: rather than proposing a new frontier model class, we study a concrete, reproducible training pipeline --- built from two existing open-weight encoders --- for the specific problem of cutting the labeling cost of a zero-shot NLI teacher for a fixed taxonomy, with results reported against measurable baselines.

Our contributions are:

\begin{enumerate}
    \item A student-guided candidate selection pipeline for efficient teacher labeling in fixed-taxonomy LLM routing, in which the first-stage model is a direct, end-to-end classifier over the true taxonomy rather than a generic embedding retriever or a clustering-derived label-tree shortlist.
    \item An iterative distillation procedure in which each improved student proposes candidates for the next teacher-labeling round, narrowing the candidate set as coverage permits.
    \item An evaluation framework that measures candidate recall, teacher agreement, classification quality, and labeling throughput together.
    \item An empirical analysis of why truncated top-$k$ teacher distributions should be separated from true full-taxonomy KL targets, connecting this failure mode to partial-label learning~\citep{cour2011partial} and to the role of non-top-class ``dark knowledge'' in soft-label distillation~\citep{hinton2015distilling}.
\end{enumerate}

\section{Related Work}
\label{sec:related-work}

This work sits at the intersection of knowledge distillation, classifier cascades, extreme multi-label classification, self-training, and LLM routing. We describe the closest lines of work in each area and state precisely where our setting differs, rather than claiming that model cascades or candidate selection are novel in general.

\paragraph{Classifier cascades and retrieve-then-rerank pipelines.} Cascade classifiers that filter easy cases with a cheap model before invoking an expensive one have a long history, dating at least to the Viola-Jones face detector~\citep{viola2001rapid}, whose cascade stages are all genuine classifiers over the same binary taxonomy (face vs.\ non-face). In modern retrieval, dense bi-encoder retrieval followed by cross-encoder reranking~\citep{karpukhin2020dpr} follows the same coarse-to-fine shape, but the first-stage model is a generic similarity/embedding function rather than a classifier trained on the same fixed label set the second stage decides over. Confident Adaptive Language Modeling~\citep{schuster2022calm} and speculative decoding~\citep{leviathan2023speculative,chen2023speculative} apply a related cheap-then-expensive idea within a single model's computation (early-exit by layer, or draft-then-verify token proposals); speculative decoding is the closest conceptual analogue in that both stages score the \emph{same} output space and the expensive stage only verifies the cheap stage's proposals, but the mechanism (parallel token verification vs.\ sequential NLI scoring of class labels) differs substantially.

\paragraph{Extreme multi-label classification (XMC).} XMC methods such as AttentionXML~\citep{you2019attentionxml}, Parabel~\citep{prabhu2018parabel}, DiSMEC~\citep{babbar2017dismec}, Slice~\citep{jain2019slice}, X-Transformer~\citep{chang2020xtransformer}, and XR-Transformer~\citep{zhang2021xrtransformer} are the closest prior \emph{architectural} pattern to our pipeline: a fast matching or shortlisting stage narrows a very large label space before an expensive ranker scores the shortlist. The distinguishing detail is how the shortlist is produced. AttentionXML and Parabel build the shortlist structure from a label tree derived by clustering label (or label-feature) embeddings, and X-Transformer's matcher typically operates over approximate-nearest-neighbor label embeddings (e.g., PIFA-style representations) rather than a direct full-taxonomy softmax. XR-Transformer instead recursively fine-tunes a single model on progressively finer label partitions rather than using two separately trained models of different capacity. In our setting the taxonomy is small (60 classes, not the $10^4$--$10^6$ label spaces XMC targets), which makes a single-pass full-distribution softmax classifier tractable as the first stage; the resulting candidate set is therefore produced by genuine end-to-end classification over the true output space, not by embedding-based approximate retrieval or by a coarsened intermediate taxonomy.

\paragraph{Distilling zero-shot NLI classifiers.} Zero-shot text classification via NLI entailment~\citep{yin2019benchmarking} is the paradigm our teacher uses, and \citet{laurer2023universal}'s work on building more efficient universal NLI classifiers (the lineage behind the \texttt{deberta-v3-large-zeroshot-v2.0} teacher we use) improves the efficiency and generality of the NLI classifier itself. That line of work, however, remains within the NLI/entailment formulation; it does not collapse zero-shot NLI scoring into a single-pass fixed-$K$-class softmax student of the kind we train. To our knowledge, no published work directly evaluates whether truncating a zero-shot NLI teacher's top-$k$ scores and zeroing the remainder produces a valid target for full-distribution KL distillation. We argue it does not: the value of soft-label distillation comes substantially from the relative probabilities assigned to \emph{non-top} classes, i.e., \citet{hinton2015distilling}'s ``dark knowledge,'' and forcing those probabilities to zero for classes the teacher never scored is not a sparse approximation of the true distribution but a systematic bias toward overconfidence, closely related to the missing-label assumption problem studied in partial-label learning~\citep{cour2011partial}. Section~\ref{sec:method} operationalizes this distinction by restricting KL supervision to fully-scored examples and using truncated top-$k$ scores only to derive hard pseudo-labels.

\paragraph{Self-training and student-guided data selection.} Self-training methods such as Noisy Student~\citep{xie2020noisystudent}, Born-Again Networks~\citep{furlanello2018born}, and weak-to-strong generalization~\citep{burns2023weaktostrong} iterate teacher$\rightarrow$student training rounds, but in each of these the teacher (or oracle) always scores the full label space for every example; the student's predictions do not narrow what the teacher is asked to evaluate. Classical active learning likewise selects which \emph{examples} receive an expensive oracle label, not which \emph{candidate labels}, for a given example, the oracle should consider. Our iterative loop differs from both families along the same axis: the student's own top-$k$ predictions prune the label space the teacher scores for that example, and this pruning tightens (smaller $k$) across iterations as measured candidate coverage permits. We are not aware of prior work that uses a student's evolving predictions to shrink an oracle's per-example label-candidate set within a fixed taxonomy in this manner.

\paragraph{LLM routing.} A separate body of work studies routing a query to one of several \emph{LLMs} to trade off cost and quality, e.g., RouteLLM~\citep{ong2024routellm}, FrugalGPT~\citep{chen2023frugalgpt}, Tryage~\citep{kwon2023tryage}, RouterBench~\citep{hu2024routerbench}, AutoMix~\citep{madaan2023automix}, MetaLLM~\citep{nguyen2024metallm}, and Hybrid LLM routing~\citep{ding2024hybrid}. These systems route between an unordered set of model choices, not into a fixed, semantically meaningful task taxonomy used to dispatch to downstream agents or tools; the routing decision itself carries no domain content. Selective classification and calibrated deferral~\citep{geifman2017selective} are methodologically relevant to any confidence-thresholded routing decision, including ours, but again are agnostic to the taxonomy-classification setting we study. To our knowledge, classifying an LLM request into a fixed task taxonomy for downstream agent routing, using an iterative teacher-guided distillation loop, has not been directly studied in prior published work.

\section{Problem Formulation}

Let:

\begin{itemize}
    \item $x$ be a user prompt;
    \item $\mathcal{Y} = \{y_1, \ldots, y_{60}\}$ be the fixed task taxonomy;
    \item $T(x, \mathcal{Y})$ be the expensive zero-shot teacher distribution;
    \item $S_t(x)$ be the student distribution at iteration $t$;
    \item $C_k(x)$ be the $k$ labels selected from the student distribution.
\end{itemize}

Full teacher labeling computes $p_T(y \mid x)$ for every $y \in \mathcal{Y}$. Student-guided labeling instead computes

\[
C_k(x) = \mathrm{TopK}(S_t(x), k), \qquad p_T(y \mid x) \text{ for } y \in C_k(x).
\]

The teacher top-1 label can only be recovered if it belongs to $C_k(x)$. We therefore define candidate coverage as

\[
\mathrm{Coverage@}k = P\big(\arg\max_y p_T(y \mid x) \in C_k(x)\big).
\]

The primary behavioral metric is teacher agreement:

\[
\mathrm{Agreement} = P\big(\arg\max_y S(y \mid x) = \arg\max_y p_T(y \mid x)\big).
\]

Human-reviewed accuracy remains necessary because teacher agreement measures imitation, not correctness.

\section{Method}
\label{sec:method}

\subsection{Teacher}

The teacher is \texttt{MoritzLaurer/deberta-v3-large-zeroshot-v2.0}~\citep{he2023debertav3,laurer2023universal}, used through the Hugging Face \texttt{zero-shot-classification} pipeline with \texttt{multi\_label=False} and the hypothesis template ``The user's task is \{\}.'' The teacher produces a top-1 label, confidence, and scores for all 60 labels when full distillation is required.

\subsection{Student}

The student is \texttt{MoritzLaurer/ModernBERT-base-zeroshot-v2.0}~\citep{warner2024modernbert} with a newly initialized 60-class classification head. For an input prompt, the student performs one encoder forward pass and emits 60 logits; softmax converts these logits into a 60-dimensional probability vector. Unlike the teacher, the student does not evaluate 60 separate prompt-label pairs; it directly predicts the complete taxonomy distribution in one forward pass.

\subsection{Distillation objective}

The student is trained with a combination of hard and soft teacher supervision:

\[
L = \alpha\, \mathrm{CE}(z_S, y_{\text{gold}}) + \gamma\, \mathrm{CE}(z_S, y_{\text{teacher}}) + \beta\, T^2\, \mathrm{KL}\big(\mathrm{softmax}(z_S / T),\, p_{\text{teacher}}^{(T)}\big)
\]

Here $z_S$ is the student logit vector, $y_{\text{teacher}}$ is the teacher top-1 label, and $p_{\text{teacher}}$ is the normalized teacher distribution over all 60 classes. For student-guided labeling, top-$k$ reranking is used to generate reliable hard labels at scale. As argued in Section~\ref{sec:related-work}, a top-$k$ score vector with the remaining classes set to zero is not a valid full 60-class teacher distribution and is not used as a KL target; it is used only to derive the hard pseudo-label $y_{\text{teacher}}$ for the $\alpha$/$\gamma$ terms above.

\subsection{Iterative procedure}

At iteration $t$:

\begin{enumerate}
    \item Run the current student on unlabeled synthetic taxonomy examples.
    \item Select top-$k$ candidate categories from the 60-way student output.
    \item Run the DeBERTa teacher only on the candidate labels.
    \item Keep labels above a confidence threshold for automatic training.
    \item Mix the new data with seed and previously reviewed data.
    \item Fine-tune the current best student checkpoint.
    \item Evaluate on a fixed, independent test set.
\end{enumerate}

The next iteration may reduce $k$ only after measuring candidate coverage: for example, top-8 can be followed by top-4 only if top-4 retains sufficient coverage of the full teacher's top-1 label.

\subsection{Deployment architecture}
\label{sec:deployment-arch}

\begin{figure}[!htb]
\centering
\begin{tikzpicture}[
  node/.style={draw, rounded corners, minimum height=0.9cm, align=center, font=\small, inner sep=6pt},
  ->, >=Latex, thick
]
\node[node] (client) {Client};
\node[node, right=3.4cm of client] (nginx) {Nginx :8080\\\texttt{least\_conn}};
\node[node, right=2.8cm of nginx, yshift=3.3cm] (v0) {vLLM pooling\\GPU 0 :: 8004\\student-v4};
\node[node, right=2.8cm of nginx, yshift=1.1cm] (v1) {vLLM pooling\\GPU 1 :: 8005\\student-v4};
\node[node, right=2.8cm of nginx, yshift=-1.1cm] (v2) {vLLM pooling\\GPU 2 :: 8006\\student-v4};
\node[node, right=2.8cm of nginx, yshift=-3.3cm] (v3) {vLLM pooling\\GPU 3 :: 8007\\student-v4};

\draw (client) -- (nginx) node[midway, above=4pt, font=\scriptsize] {POST /router/classify};
\draw (nginx) -- (v0);
\draw (nginx) -- (v1);
\draw (nginx) -- (v2);
\draw (nginx) -- (v3);
\end{tikzpicture}
\caption{Production serving architecture for the deployed v4 student. Nginx load-balances (\texttt{least\_conn}) across four independent vLLM pooling-runner instances, one per GPU (4$\times$ NVIDIA RTX PRO 6000 Blackwell), each running a full replica of the same student checkpoint. This is the architecture measured in the server-local throughput benchmark of Section~\ref{sec:prelim-measurements} ($\sim$5,200--5,250 req/s).}
\label{fig:deployment}
\end{figure}
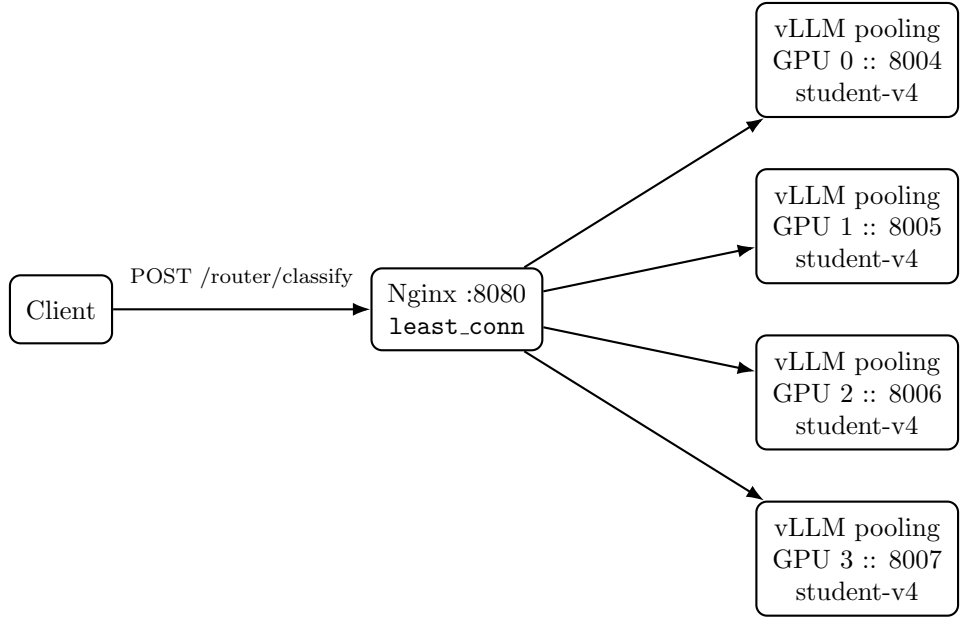

The trained student is served in production behind Nginx, which load-balances across four independent vLLM pooling-runner instances, one per GPU, as shown in Figure~\ref{fig:deployment}. This is the configuration used for the live-endpoint measurements in Section~\ref{sec:prelim-measurements}: a client issues \texttt{POST /router/classify} to Nginx on port 8080, which distributes requests with a \texttt{least\_conn} policy across four \texttt{vllm serve} processes (ports 8004--8007), each holding an independent copy of the same student checkpoint on its own GPU with \texttt{--gpu-memory-utilization 0.20} and \texttt{--max-model-len 512}. Because each instance is a full replica rather than a shard, throughput scales close to additively with instance count (Section~\ref{sec:prelim-measurements}), at the cost of $4\times$ the GPU memory a single instance would use.

\section{Data}

The taxonomy contains 60 task categories derived from an internal \texttt{task\_request\_mapping.json} schema. Experiments use seed examples from the fixed taxonomy, \texttt{HuggingFaceH4/no\_robots} requests, \texttt{PrimeIntellect/SWE-rebench-V2} software-engineering requests, and an internal synthetic taxonomy dataset of LLM-generated, human-reviewed golden examples covering the 60 categories, used in sizes ranging from 1,000 examples in early experiments up to 2,218 examples in later full-60 teacher-scored training runs (Table~\ref{tab:versions}). All training and evaluation data is drawn from public datasets or synthetically generated, human-reviewed examples; no real user traffic or production request records were accessed or used at any stage.

The synthetic taxonomy dataset was generated by prompting an LLM to produce example user requests for each of the 60 categories, followed by manual human review to confirm category fidelity and remove low-quality or duplicate examples. Prompts longer than 8,000 characters are excluded before labeling.

\section{Experimental Results So Far}

\subsection{Student versions}

\begin{table}[h]
\centering
\begin{tabular}{@{}llr@{}}
\toprule
Version & Training data / objective & Teacher agreement \\
\midrule
Baseline & 300 seed examples + 221 high-confidence pseudo-labels & Early baseline \\
v2 & Public SWE teacher soft labels & 62.82\% \\
v3 & Public SWE + 1{,}000 synthetic taxonomy examples & 69.5\% (139/200) \\
v4 & Teacher-focused distillation, initialized from ModernBERT-base & 76.5\% (153/200) \\
v5 & Public SWE + top-8 synthetic-taxonomy labels & 75.0\% (150/200) \\
v5-full60 & 2{,}218 full-60 teacher-scored synthetic samples, init.\ from v4 & 71.5\% (143/200) \\
v5-tuned & Same 2{,}218 full-60 samples, init.\ from v4, 1 epoch, lr 5e-6 & \textbf{77.5\% (155/200)} \\
\bottomrule
\end{tabular}
\caption{Student checkpoints evaluated so far. v5-tuned is the best checkpoint measured to date but has not been promoted to production; v4 remains the currently deployed checkpoint and is the one used for the live-endpoint measurements in Section~\ref{sec:prelim-measurements}. These results are preliminary and use a 200-example evaluation set; they do not establish production-level accuracy and are reported here as intermediate progress, not as a final benchmark.}
\label{tab:versions}
\end{table}

\subsection{Observations}

The top-$k$ labeling pipeline substantially increases teacher-labeling throughput compared with evaluating all 60 labels for every request. However, adding more synthetic data did not automatically improve student agreement: v5-full60 underperforms v4 despite using an order of magnitude more teacher-scored examples, and only a further short, low-learning-rate fine-tune from v4 (v5-tuned) narrowly surpassed v4 to become the best checkpoint measured so far. The main observed risks are low-confidence pseudo-labels, duplicated examples, distribution shift between public SWE data and the synthetic taxonomy data, and invalid truncated soft targets used as KL supervision in earlier training runs.

\subsection{Preliminary coverage, agreement, and latency measurements}
\label{sec:prelim-measurements}

We report a first pass at candidate coverage@$k$, top-$k$ teacher-rerank agreement, and end-to-end latency for the deployed v4 student, measured against its live inference endpoint. (v5-tuned, at 77.5\% teacher agreement, is the best checkpoint measured to date, but it had not been promoted to the production endpoint at measurement time, so the live-endpoint numbers below characterize v4, not v5-tuned.) These numbers partially address items~2--4 of Section~\ref{sec:required-eval} but fall short of the full required evaluation in two ways that we state explicitly: neither sample below is the certified, independently held-out 200-example set behind the agreement figures in Table~\ref{tab:versions} (the training host holding that set was unreachable at measurement time), and the per-request latency figures for the two agreement measurements below are measured over a corporate WAN link to the deployment host rather than from a co-located, in-cluster client (we do separately report an in-cluster server-local throughput number below).

\textbf{Seed sanity check (in-sample, $n=300$).} We used \texttt{data/teacher\_seed.jsonl}, the 300-example hand-authored seed set with pre-computed full-60 teacher scores. Because this set was very likely part of v4's own training data, results are an in-sample plausibility check, not an independent measurement: student--teacher top-1 agreement was only 39.0\%, and student top-1 vs.\ the human gold label was 19.0\%, both far below v4's 76.5\% headline figure. We attribute this gap primarily to distribution shift rather than a labeling or evaluation bug (we verified the label vocabularies match exactly): per Table~\ref{tab:versions}, v4 was trained on public SWE data plus synthetic taxonomy samples, and the hand-authored seed set's phrasing may differ enough from that training distribution to depress agreement substantially, which is itself a relevant finding about the pipeline's sensitivity to training-serving skew. Despite low top-1 agreement, candidate coverage was comparatively strong: Coverage@4~=~74.3\%, Coverage@8~=~84.3\%, Coverage@16~=~91.3\%, with top-$k$-rerank agreement equal to coverage at each $k$ (consistent with the softmax-argmax invariance used to derive it; see Appendix discussion in the released scripts). WAN latency for a single classify call was low: mean 158\,ms, p50 156\,ms, p95 167\,ms, p99 172\,ms.

\textbf{Fresh SWE-rebench holdout (best-effort, $n=40$).} To reduce (but not eliminate) in-sample bias, we drew 40 fresh examples from \texttt{PrimeIntellect/SWE-rebench-V2} starting at row offset 4{,}000 --- past the row range most likely used in prior training pulls, though the exact training row range is not recorded in this checkout, so this is a best-effort held-out set rather than a certified-disjoint one. We scored each example with a freshly loaded copy of the real DeBERTa-v3-large teacher (run locally on CPU) over all 60 labels, and queried the live v4 endpoint for the student distribution. No human gold label exists for these examples, so we report only teacher-imitation metrics. Student--teacher top-1 agreement was 62.5\%, closer to v4's own 76.5\% headline figure than the seed check above. Coverage@$k$ was Coverage@4~=~80.0\%, Coverage@8~=~92.5\%, Coverage@16~=~100\%, again with top-$k$-rerank agreement matching coverage exactly at each $k$. WAN latency for the student endpoint was similar to the seed check: mean 160\,ms, p50 159\,ms, p95 169\,ms.

\textbf{Latency and throughput sweep (WAN).} We additionally load-tested the live student endpoint with concurrent \texttt{aiohttp} keep-alive clients across a $3\times3$ grid of request batch size $\in \{1, 8, 32\}$ and client concurrency $\in \{1, 8, 32\}$, 15 seconds per configuration. At batch size 1 and concurrency 1, p50 latency was 76.4\,ms at 12.9 requests/s. At batch size 32 and concurrency 32, throughput reached 242.6 requests/s (7{,}763.2 classifications/s) at p50 111.5\,ms, with 7 failed requests out of 3{,}646 at that highest-load configuration. This is a WAN benchmark from a single external client, not an in-cluster, pod-to-pod load test, and should not be read as a server capacity ceiling.

\textbf{Server-local throughput (in-cluster).} To obtain a server-capacity number uncontaminated by WAN round-trip time and client-side connection overhead, we separately benchmarked the deployed v4 router with \texttt{wrk} running directly on the deployment host (\texttt{g4-4}), issuing single-example, non-batched requests over persistent HTTP connections to all four load-balanced vLLM pooling instances (one per GPU, four NVIDIA RTX PRO 6000 Blackwell cards) behind the same Nginx \texttt{least\_conn} endpoint used above. At 4 threads and 64 connections, server-local \texttt{wrk} throughput was stable at approximately 5{,}200--5{,}250 requests/s with 12.15--12.29\,ms average latency; each individual vLLM instance alone sustained roughly 1{,}290--1{,}427 requests/s at 64 concurrent connections (p50 34--38\,ms, p95 44--51\,ms), confirming the four instances scale close to additively under Nginx load balancing. For comparison, \texttt{ApacheBench} against the same server-local endpoint measured only 2{,}174--2{,}185 requests/s, and a remote \texttt{ApacheBench} client (256 connections, non-persistent) measured just 329.53 requests/s at 776.87\,ms average latency --- both artifacts of each tool's own connection model and (for the remote case) network path, not evidence of the server's real capacity. This single-example, in-cluster figure is not directly comparable to the WAN batch-of-32 throughput above, since the two measure different request shapes (unbatched vs.\ client-side batched inputs) and different vantage points (local vs.\ WAN); together, they bound the deployed router's practical throughput between roughly 5,200 single-example requests/s at the server and 150--170\,ms end-to-end latency for a single request from an external client.

\textbf{Reading these results together.} Coverage@16 reaches 91--100\% in both samples, and coverage degrades gracefully rather than collapsing at smaller $k$, which is consistent with (but does not yet prove) the paper's central claim that student-guided top-$k$ selection preserves most of the teacher's decision-relevant information at a fraction of the full 60-way labeling cost. The large gap between the in-sample seed check (39\% agreement) and the fresh holdout (62.5\% agreement) shows that agreement is highly sensitive to how close the evaluation distribution is to the student's training distribution, reinforcing why Section~\ref{sec:required-eval} calls for a larger, explicitly held-out, human-reviewed evaluation set rather than relying on either of these two convenience samples as a headline number. The server-local throughput measurement of ~5,200 requests/s per four-GPU deployment is the first concrete evidence toward the paper's motivating cost argument, though it still needs to be paired with an equivalent full-60-teacher, in-cluster GPU throughput number (not yet measured; our only full-60 teacher timings are unrepresentative local-CPU runs, see Section~\ref{sec:required-eval}) before a labeling-cost-reduction ratio can be reported.

\section{Required Evaluation}
\label{sec:required-eval}

Substantiating the pipeline's central claim requires comparing student-guided candidate selection at several values of $k$ against a full-teacher upper bound, an embedding-retrieval baseline, and a student-only lower bound, on a properly held-out, human-reviewed sample, reporting candidate recall, agreement, accuracy, latency, and cost for each. The complete comparison matrix and metric list are tracked in the released repository's \texttt{experiment-punch-list.md} rather than reproduced here, since that plan is still being executed.

\section{Reproducibility and Release Plan}
\label{sec:reproducibility}

The repository contains scripts for seed preparation, teacher labeling, synthetic data generation, prompt filtering, student-guided retrieval, student training, and evaluation. Model artifacts remain outside the public repository. A reproducible release will include the fixed label schema and category descriptions; synthetic or de-identified example prompts; model and tokenizer identifiers and revisions; all loss and inference hyperparameters; train/validation/test split hashes; candidate coverage and agreement reports; and throughput and hardware measurements.

\section{Conclusion}

This paper presents a student-guided teacher distillation pipeline for efficient 60-class LLM task routing. The student is used not only as the final fast router, but also as a task-aware candidate generator that reduces expensive zero-shot teacher labeling, distinguishing it from generic embedding-based candidate generation used in extreme multi-label classification cascades and from LLM-choice routing methods that do not classify into a semantic task taxonomy at all. Preliminary experiments demonstrate a best teacher agreement of 77.5\% (v5-tuned; the currently deployed v4 checkpoint reaches 76.5\%), while also identifying important failure modes around low-confidence data and truncated soft targets, which we connect to the theoretical grounds for why partial teacher observations should not be treated as full-distribution KL targets. A first, best-effort measurement of candidate coverage (Section~\ref{sec:prelim-measurements}), run against the deployed v4 checkpoint, shows Coverage@16 of 91--100\% across two convenience samples, and highlights that agreement is sensitive to distribution shift between evaluation and training data --- a factor the certified held-out evaluation in Section~\ref{sec:required-eval} still needs to control for.

The central empirical question for the completed paper is whether iterative student-guided candidate selection achieves a better accuracy--labeling-cost trade-off than full teacher labeling, embedding retrieval, and non-iterative distillation. Section~\ref{sec:required-eval} scopes the specific measurements --- candidate coverage across $k$, an embedding-retrieval baseline, and a larger human-reviewed evaluation set --- required to answer it, and we intend to report them in a subsequent revision of this paper.

\bibliographystyle{unsrtnat}
\bibliography{references}

@article{warner2024modernbert,
  title={Smarter, Better, Faster, Longer: A Modern Bidirectional Encoder for Fast, Memory Efficient, and Long Context Finetuning and Inference},
  author={Warner, Benjamin and Chaffin, Antoine and Clavi{\'e}, Benjamin and Weller, Orion and Hallstr{\"o}m, Oskar and Taghadouini, Said and Gallagher, Alexis and Biswas, Raja and Ladhak, Faisal and Aarsen, Tom and Cooper, Nathan and Adams, Griffin and Howard, Jeremy and Poli, Iacopo},
  journal={arXiv preprint arXiv:2412.13663},
  year={2024}
}

@article{he2023debertav3,
  title={DeBERTaV3: Improving DeBERTa using ELECTRA-Style Pre-Training with Gradient-Disentangled Embedding Sharing},
  author={He, Pengcheng and Gao, Jianfeng and Chen, Weizhu},
  journal={arXiv preprint arXiv:2111.09543},
  year={2021}
}

@article{hinton2015distilling,
  title={Distilling the Knowledge in a Neural Network},
  author={Hinton, Geoffrey and Vinyals, Oriol and Dean, Jeff},
  journal={arXiv preprint arXiv:1503.02531},
  year={2015}
}

@article{laurer2023universal,
  title={Building Efficient Universal Classifiers with Natural Language Inference},
  author={Laurer, Moritz and van Atteveldt, Wouter and Casas, Andreu and Welbers, Kasper},
  journal={arXiv preprint arXiv:2312.17543},
  year={2023}
}

@article{yin2019benchmarking,
  title={Benchmarking Zero-shot Text Classification: Datasets, Evaluations and Entailment Approach},
  author={Yin, Wenpeng and Hay, Jamaal and Roth, Dan},
  journal={arXiv preprint arXiv:1909.00161},
  year={2019}
}

@article{you2019attentionxml,
  title={AttentionXML: Label Tree-based Attention-Aware Deep Model for High-Performance Extreme Multi-Label Text Classification},
  author={You, Ronghui and Zhang, Zihan and Wang, Ziye and Dai, Suyang and Mamitsuka, Hiroshi and Zhu, Shanfeng},
  journal={arXiv preprint arXiv:1811.01727},
  year={2019}
}

@article{zhang2021xrtransformer,
  title={Fast Multi-Resolution Transformer Fine-tuning for Extreme Multi-label Text Classification},
  author={Zhang, Jiong and Chang, Wei-Cheng and Yu, Hsiang-Fu and Dhillon, Inderjit S.},
  journal={arXiv preprint arXiv:2110.00685},
  year={2021}
}

@article{chang2020xtransformer,
  title={Taming Pretrained Transformers for Extreme Multi-label Text Classification},
  author={Chang, Wei-Cheng and Yu, Hsiang-Fu and Zhong, Kai and Yang, Yiming and Dhillon, Inderjit S.},
  journal={arXiv preprint arXiv:1905.02331},
  year={2020}
}

@inproceedings{prabhu2018parabel,
  title={Parabel: Partitioned Label Trees for Extreme Classification with Application to Dynamic Search Advertising},
  author={Prabhu, Yashoteja and Kag, Anil and Harsola, Shrutendra and Agrawal, Rahul and Varma, Manik},
  booktitle={Proceedings of the 2018 World Wide Web Conference (WWW)},
  year={2018}
}

@article{babbar2017dismec,
  title={DiSMEC: Distributed Sparse Machines for Extreme Multi-label Classification},
  author={Babbar, Rohit and Sch{\"o}lkopf, Bernhard},
  journal={arXiv preprint arXiv:1609.02521},
  year={2017}
}

@inproceedings{jain2019slice,
  title={Slice: Scalable Linear Extreme Classifiers Trained on 100 Million Labels for Related Searches},
  author={Jain, Himanshu and Balasubramanian, Venkatesh and Chunduri, Bhanu and Varma, Manik},
  booktitle={Proceedings of the Twelfth ACM International Conference on Web Search and Data Mining (WSDM)},
  year={2019}
}

@article{chen2023frugalgpt,
  title={FrugalGPT: How to Use Large Language Models While Reducing Cost and Improving Performance},
  author={Chen, Lingjiao and Zaharia, Matei and Zou, James},
  journal={arXiv preprint arXiv:2305.05176},
  year={2023}
}

@article{ong2024routellm,
  title={RouteLLM: Learning to Route LLMs with Preference Data},
  author={Ong, Isaac and Almahairi, Amjad and Wu, Vincent and Chiang, Wei-Lin and Wu, Tianhao and Gonzalez, Joseph E. and Kadous, M. Waleed and Stoica, Ion},
  journal={arXiv preprint arXiv:2406.18665},
  year={2024}
}

@article{leviathan2023speculative,
  title={Fast Inference from Transformers via Speculative Decoding},
  author={Leviathan, Yaniv and Kalman, Matan and Matias, Yossi},
  journal={arXiv preprint arXiv:2211.17192},
  year={2023}
}

@article{chen2023speculative,
  title={Accelerating Large Language Model Decoding with Speculative Sampling},
  author={Chen, Charlie and Borgeaud, Sebastian and Irving, Geoffrey and Lespiau, Jean-Baptiste and Sifre, Laurent and Jumper, John},
  journal={arXiv preprint arXiv:2302.01318},
  year={2023}
}

@inproceedings{viola2001rapid,
  title={Rapid Object Detection using a Boosted Cascade of Simple Features},
  author={Viola, Paul and Jones, Michael},
  booktitle={Proceedings of the 2001 IEEE Computer Society Conference on Computer Vision and Pattern Recognition (CVPR)},
  year={2001}
}

@article{schuster2022calm,
  title={Confident Adaptive Language Modeling},
  author={Schuster, Tal and Fisch, Adam and Gupta, Jai and Dehghani, Mostafa and Bahri, Dara and Tran, Vinh Q. and Tay, Yi and Metzler, Donald},
  journal={arXiv preprint arXiv:2207.07061},
  year={2022}
}

@article{karpukhin2020dpr,
  title={Dense Passage Retrieval for Open-Domain Question Answering},
  author={Karpukhin, Vladimir and O{\u{g}}uz, Barlas and Min, Sewon and Lewis, Patrick and Wu, Ledell and Edunov, Sergey and Chen, Danqi and Yih, Wen-tau},
  journal={arXiv preprint arXiv:2004.04906},
  year={2020}
}

@article{kwon2023tryage,
  title={Tryage: Real-time, intelligent Routing of User Prompts to Large Language Models},
  author={Kwon, Surin and Almalis, Chaithra and Daniel, Elias and Kang, Emma and Song, Sun and Gee, James and Vidal, Jorge Mario},
  journal={arXiv preprint arXiv:2308.11601},
  year={2023}
}

@article{hu2024routerbench,
  title={RouterBench: A Benchmark for Multi-LLM Routing System},
  author={Hu, Qitian Jason and Bieker, Jacob and Li, Xiuyu and Jiang, Nan and Keigwin, Benjamin and Ranganath, Gaurav and Keutzer, Kurt and Upadhyay, Shriyash Kaustubh},
  journal={arXiv preprint arXiv:2403.12031},
  year={2024}
}

@article{madaan2023automix,
  title={AutoMix: Automatically Mixing Language Models},
  author={Madaan, Aman and Aggarwal, Pranjal and Anand, Ankit and Potharaju, Srividya Pranavi and Mishra, Swaroop and Zhou, Pei and Gupta, Aditya and Rajagopal, Dheeraj and Kappaganthu, Karthik and Yang, Yiming and Upadhyay, Shyam and Faruqui, Manaal},
  journal={arXiv preprint arXiv:2310.12963},
  year={2023}
}

@article{nguyen2024metallm,
  title={MetaLLM: A High-performant and Cost-efficient Dynamic Framework for Wrapping LLMs},
  author={Nguyen, Quang H. and Hoang, Duy C. and Decugis, Juliette and Manchanda, Saurav and Chawla, Nitesh V. and Doan, Khoa D.},
  journal={arXiv preprint arXiv:2407.10834},
  year={2024}
}

@article{ding2024hybrid,
  title={Hybrid LLM: Cost-Efficient and Quality-Aware Query Routing},
  author={Ding, Dujian and Mallick, Ankur and Wang, Chi and Sim, Robert and Mukherjee, Subhabrata and Ruhle, Victor and Lakshmanan, Laks V. S. and Awadallah, Ahmed Hassan},
  journal={arXiv preprint arXiv:2404.14618},
  year={2024}
}

@article{geifman2017selective,
  title={Selective Classification for Deep Neural Networks},
  author={Geifman, Yonatan and El-Yaniv, Ran},
  journal={arXiv preprint arXiv:1705.08500},
  year={2017}
}

@article{yu2024distilling,
  title={Distilling System 2 into System 1},
  author={Yu, Ping and Xu, Jing and Weston, Jason and Kulikov, Ilia},
  journal={arXiv preprint arXiv:2407.06023},
  year={2024}
}

@article{xie2020noisystudent,
  title={Self-training with Noisy Student improves ImageNet classification},
  author={Xie, Qizhe and Luong, Minh-Thang and Hovy, Eduard and Le, Quoc V.},
  journal={arXiv preprint arXiv:1911.04252},
  year={2020}
}

@article{burns2023weaktostrong,
  title={Weak-to-Strong Generalization: Eliciting Strong Capabilities With Weak Supervision},
  author={Burns, Collin and Izmailov, Pavel and Kirchner, Jan Hendrik and Baker, Bowen and Gao, Leo and Aschenbrenner, Leopold and Chen, Yining and Ecoffet, Adrien and Joglekar, Manas and Leike, Jan and Sutskever, Ilya and Wu, Jeff},
  journal={arXiv preprint arXiv:2312.09390},
  year={2023}
}

@article{furlanello2018born,
  title={Born Again Neural Networks},
  author={Furlanello, Tommaso and Lipton, Zachary C. and Tschannen, Michael and Itti, Laurent and Anandkumar, Anima},
  journal={arXiv preprint arXiv:1805.04770},
  year={2018}
}

@misc{typesafe2026jev,
  title={Introducing System One Models and Jev},
  author={{TypeSafe AI}},
  year={2026},
  howpublished={\url{https://typesafe.ai/blog/introducing-system-one-models-and-jev}},
  note={Blog post; no accompanying technical report available at time of writing}
}

@article{yang2023rlcd,
  title={RLCD: Reinforcement Learning from Contrastive Distillation for Language Model Alignment},
  author={Yang, Kevin and Klein, Dan and Celikyilmaz, Asli and Peng, Nanyun and Tian, Yuandong},
  journal={arXiv preprint arXiv:2307.12950},
  year={2023},
  note={Unrelated to TypeSafe AI's use of the acronym RLCD; name collision only}
}

@misc{laya2026,
  title={Laya: Multilingual, Non-Autoregressive System 1 Decision Engine},
  author={{Convai Innovations} and {NandhaKishorM}},
  year={2026},
  howpublished={\url{https://github.com/NandhaKishorM/laya}},
  note={GitHub repository, Apache 2.0 license; no accompanying technical report or peer-reviewed publication at time of writing}
}

@article{cour2011partial,
  title={Learning from Partial Labels},
  author={Cour, Timothee and Sapp, Ben and Taskar, Ben},
  journal={Journal of Machine Learning Research},
  volume={12},
  pages={1501--1536},
  year={2011}
}

\end{document}